\documentclass[11pt,a4paper]{article}
\usepackage[table,xcdraw,dvipsnames]{xcolor}
\usepackage{times,latexsym,linguex}
\usepackage{url,subcaption}
\usepackage{babel}
\usepackage{caption}
\usepackage[normalem]{ulem}
\usepackage[T1]{fontenc}

\usepackage[acceptedWithA]{tacl2018v2}

\usepackage{xspace,mfirstuc,tabulary}

\newif\iftaclinstructions
\taclinstructionsfalse %
\iftaclinstructions
\renewcommand{\confidential}{}
\renewcommand{\anonsubtext}{(No author info supplied here, for consistency with
TACL-submission anonymization requirements)}
\newcommand{\instr}
\fi

\iftaclpubformat %

\else

\fi

\usepackage{url}
\usepackage{latexsym}
\usepackage{booktabs}
\usepackage{stmaryrd}
\usepackage{amsmath}
\usepackage{multirow}
\usepackage{caption}
\usepackage{babel}
\usepackage{textcomp}
\usepackage{todonotes}

\setuptodonotes{color=blue!30}

\newcommand{\M}{$^M$}
\newcommand{\event}[2]{
  \uline{#1}$^{\text{#2}}$
}
\newcommand{\eventp}[3]{
  \uline{#1}$^{\text{#2}\  [\text{#3}]}$
}

\newcommand{\cev}[1]{\textcolor{NavyBlue}{\textbf{#1}}}
\newcommand{\env}[1]{\textcolor{NavyBlue}{\textsl{#1}}}

\title{He Thinks He Knows Better than the Doctors: \\
  BERT for Event Factuality Fails on Pragmatics}

\author{Nanjiang Jiang \\
Department of Linguistics \\
The Ohio State University \\
  {\tt jiang.1879@osu.edu} \\\And
  Marie-Catherine de Marneffe\\
Department of Linguistics \\
The Ohio State University \\
  {\tt demarneffe.1@osu.edu} \\}

\date{}

\begin{document}
\setlength{\Exlabelsep}{.5pt}
\setlength{\Exlabelwidth}{2em}
\setlength{\Extopsep}{0.4\baselineskip}

\maketitle

\begin{abstract}
We investigate how well BERT performs on predicting factuality in several existing English datasets, encompassing various linguistic constructions. Although BERT obtains a strong performance on most datasets, it does so by exploiting common surface patterns that correlate with certain factuality labels, and it fails on instances where pragmatic reasoning is necessary. Contrary to what the high performance suggests, we are still far from having a robust system for factuality prediction.
\end{abstract}

\section{Introduction}

\begin{table*}[t!]
    \centering
    \resizebox{\textwidth}{!}{
    \begin{tabular}{l|p{18cm}}
\toprule
      MegaVeridicality &
                          Someone was \cev{misinformed} that something \event{happened}{-2.7}. \\
      CB &
                       Hazel had not felt so much bewildered since Blackberry had talked about the raft beside the Enborne. Obviously, the stones could not possibly be anything to do with El-ahrairah. It seemed to him that Strawberry \env{might} as well have \cev{said} that his tail \event{was}{-1.33} an oak tree.\\
      RP &
           The man \cev{managed} to \event{stay}{3} on his horse.	/
           The man did \env{not} \cev{manage} to \event{stay}{-2.5} on his horse.	\\
\midrule
FactBank &
Helicopters are \event{flying}{3.0} over northern New York today \event{\cev{trying}}{3.0} to
\event{locate}{0} people \event{stranded}{3.0} without food, heat or medicine.\\
MEANTIME &
Alongside both \event{announcements}{3.0}, Jobs also \event{announced}{3.0} a new iCloud service to \event{sync}{0} data among all devices.\\
UW &
Those plates may have \event{come}{1.4} from a machine shop in north Carolina, where a friend of Rudolph \event{worked}{3.0}.\\
UDS-IH2 &
DPA: Iraqi authorities \event{\cev{announced}}{2.25} that they had \event{busted}{2.625} up 3 terrorist cells \event{operating}{2.625} in Baghdad. \\
\bottomrule
    \end{tabular}
    }
    \caption{Example items from each dataset. The annotated event predicates are underlined with their factuality annotations in superscript.
    For the datasets focusing on embedded events (first group), the
    clause-embedding verbs are in bold and the entailment-canceling environments
    (if any) are slanted.
    }
    \label{tab:examples}
\end{table*}

Predicting event factuality\footnote{The terms \textit{veridicality} and \textit{speaker
  commitment} refer to the same underlying linguistic phenomenon.} is the task of identifying to what extent
an event mentioned in a sentence is presented by the author as factual.
It is a complex semantic and pragmatic phenomenon:
in \textit{John thinks he knows better than the doctors}, we infer that John
probably doesn't know better than the doctors.
Event factuality inference is prevalent in human communication and matters for
tasks which depend on natural language understanding, such as information
extraction. For instance, in the FactBank example \citep{sauri2009factbank}  in Table~\ref{tab:examples}, an information extraction system should extract \textit{people are stranded without food} but not \textit{helicopters located people stranded without food}.

The current state-of-the-art model for factuality prediction on English is  \citet{pouran-ben-veyseh-etal-2019-graph}, obtaining the best performance on four factuality datasets:
FactBank, MEANTIME
\citep{minard2016meantime}, UW \cite{lee2015event} and UDS-IH2 \cite{rudinger}.
Traditionally, event factuality is thought to be triggered by fixed properties of lexical items. The Rule-based model of \citet{stanovsky} took such an approach:
they used lexical rules and dependency trees to determine whether an event in a
sentence is factual, based on the properties of the lexical items that embed the event in question.
\citet{rudinger} proposed the first end-to-end model for factuality with LSTMs. \citet{pouran-ben-veyseh-etal-2019-graph} used BERT representations with a graph convolutional network and obtained a large improvement over \citet{rudinger} and over \citet{stanovsky}'s Rule-based
model (except for one metric on the UW dataset).

However, it is not clear what these end-to-end models learn and what
features are encoded in their representations.
In particular, they do not seem capable of generalizing to events embedded under certain
linguistic constructions. \citet{white-etal-2018-lexicosyntactic} showed that the \citet{rudinger}'s models
exhibit systematic errors on MegaVeridicality, which contains
factuality inferences purely triggered by the semantics of clause-embedding verbs in specific syntactic contexts.
\citet{jiang-de-marneffe-2019-know} showed that \citeauthor{stanovsky}'s and
\citeauthor{rudinger}'s models fail
to perform well on the CommitmentBank \citep{demarneffe2018} which contains
events under clause-embedding verbs in an entailment-canceling environment (negation, question, modal or antecedent of conditional).

In this paper, we investigate how well BERT, using a standard fine-tuning
approach,\footnote{We only augment BERT with a task-specific layer, instead of
  proposing a new task-specific model as in \citet{pouran-ben-veyseh-etal-2019-graph}.}
performs on seven factuality datasets,
including those focusing on embedded events which
have been shown to be challenging (\citealt{white-etal-2018-lexicosyntactic} and
\citealt{jiang-de-marneffe-2019-know}). The application of BERT to datasets focusing on embedded events
has been limited to the setup of natural language inference (NLI) \cite{poliak-etal-2018-collecting-diverse, jiang-de-marneffe-2019-evaluating, ross-pavlick-2019-well}.
In the NLI setup, an item is a premise-hypothesis pair, with a categorical label for whether the event described in the hypothesis can be inferred by the premise. The categorical labels are obtained by discretizing the original real-valued annotations.
For example, given the premise \textit{the man
  managed to stay on his horse} (RP example in Table~\ref{tab:examples}) and the hypothesis \textit{the man stayed on his
  horse}, a model should predict that the hypothesis can be inferred from the premise. In the factuality setup,
an item contains a sentence with one or more spans corresponding to events, with real-valued annotations for the factuality of the event. By adopting the event factuality setup, we study whether models can predict not only the polarity but also the gradience in factuality judgments (which is removed in the NLI-style discretized labels).
Here, we provide an in-depth analysis to understand which kind of items BERT fares well on, and which kind it fails on. Our  analysis shows that, while BERT can pick up on subtle surface patterns, it consistently fails on items where the surface patterns do not lead to the factuality labels frequently associated with the pattern, and for which pragmatic reasoning is necessary.

\section{Event factuality datasets}
Several event factuality datasets for English have been introduced, with
examples from each shown in Table~\ref{tab:examples}. These datasets differ with respect to some of the features that affect event factuality.

\paragraph{Embedded events}
The datasets differ with respect to which events are annotated for factuality.
The first category, including MegaVeridicality
\citep{white-etal-2018-lexicosyntactic}, CommitmentBank (CB), and \citet{ross-pavlick-2019-well} (RP),
only contains sentences with clause-embedding verbs and factuality is  annotated solely for the
event described by the embedded clause.
These datasets were used to study speaker commitment towards the embedded content,
evaluating theories of lexical semantics
\citep[i.a.,][]{kiparsky1970,10.2307/412084,Beaver:2010:LINT}, and probing
whether neural model representations contain lexical semantic information.
In the datasets of the second category (FactBank, MEANTIME, UW and UDS-IH2),
events in both main clauses and embedded clauses (if any) are annotated.
For instance, the example for UDS-IH2 in Table~\ref{tab:examples} has annotations for the main clause event \textit{announced} and the embedded clause event \textit{busted}, while the example
for RP is annotated only for the embedded clause event \textit{stay}, but not for the main clause event \textit{managed}.

\paragraph{Genres}
The datasets also differ in genre: FactBank, MEANTIME and UW are newswire data.
Since newswire sentences tend to describe factual events, these datasets have annotations biased towards factual.
UDS-IH2, an extension of \citet{white-etal-2016-universal}, comes from the English Web Treebank \citep{bies2012english} containing weblogs, emails, and other web text.
CB comes from three genres: newswire (Wall Street Journal), fiction (British National Corpus), and dialog (Switchboard).
RP contains short sentences sampled from MultiNLI \citep{N18-1101} from 10 different genres.
MegaVeridicality contains artificially constructed ``semantically bleached'' sentences to remove confound
of pragmatics and world-knowledge, and to collect baseline judgments of how much the
verb by itself affects the factuality of the content of its complement in certain syntactic constructions.

\paragraph{Entailment-canceling environments}
The three datasets in the first category differ with respect to whether the
clause-embedding verbs are under some entailment-canceling environment, such as negation.
Under the framework of implicative signatures
\citep{10.2307/412084,nairn-etal-2006-computing,karttunen2012}, a
clause-embedding verb (in a certain syntactic frame, details later) has a
lexical semantics (a signature) indicating whether the content of its complement
is factual (\texttt{+}), nonfactual (\texttt{-}), or neutral (\texttt{o}, no indication of whether the event is factual or not).
A verb signature has the form \texttt{X/Y}, where \texttt{X} is the factuality of the content of the clausal complement when the sentence has positive polarity (not embedded under any
entailment-canceling environment), and \texttt{Y} is
the factuality when the clause-embedding verb is under negation. In the
RP example in Table~\ref{tab:examples}, \textit{manage to} has signature \texttt{+/-} which, in the positive polarity sentence \textit{the man managed to stay on his horse}, predicts the embedded event \textit{stay} to be factual (such intuition is corroborated by the $+3$ human annotation). Conversely, in the negative polarity sentence \textit{the man did not manage to
  stay on his horse}, the \texttt{-} signature signals that \textit{stay} is nonfactual (again corroborated by the $-2.5$ human annotation). For \textit{manage to}, negation cancels the factuality of its embedded event.

While such a framework assumes that different entailment-canceling environments (negation, modal, question, and antecedent of conditional) have the same effects on the factuality of the content of the complement \citep{chierchia1990meaning}, there is evidence for varying effects of environments.
\citet{doi:10.1080/08351817109370248} points out that, while the content of complement
of verbs such as \textit{realize} and \textit{discover} stays factual under negation (compare \ref{ex:pos} and \ref{ex:neg}), it does not under a question \ref{ex:question} or in the antecedent of a conditional \ref{ex:cond}.

{\small
\ex. \label{ex:pos} I \cev{realized} that
\uline{I had not told the truth}.$^\texttt{+}$

\ex. \label{ex:neg} I \env{didn't} \cev{realize} that
\uline{I had not told the truth}.$^\texttt{+}$

\ex. \label{ex:question} \env{Did} you \cev{realize} that
\uline{you had not told the truth}$^\texttt{o}$\env{?}

\ex.\label{ex:cond} \env{If} I \cev{realize} later that
\uline{I have not told the truth}$^\texttt{o}$, I will confess it to everyone.

}

\citet{smith2014relationship} provided experimental evidence that the content of the complement
of \textit{know} is perceived as more factual when \textit{know} is under negation than when it is in the antecedent of a conditional.

In MegaVeridicality, each positive polarity sentence is paired with a
negative polarity sentence where the clause-embedding verb is negated.
Similarly in RP, for each naturally occurring sentence of positive polarity,
a minimal pair negative polarity sentence was automatically generated.
The verbs in CB appear in four entailment-canceling environments:
negation, modal, question, and antecedent of conditional.

\paragraph{Frames}
Among the datasets in the first category,
the clause-embedding verbs are under different syntactic contexts/frames, which also
affect the factuality of their embedded events.
For example, \textit{forget}
has signature \texttt{+/+} in \textit{forget that S}, but \texttt{-/+}
in \textit{forget to VP}.
That is, in \textit{forget that S}, the content of the clausal
complement \textit{S} is factual in both \textit{someone forgot that S} and
\textit{someone didn't forget that S}.
In \textit{forget to VP}, the content of the infinitival complement
\textit{VP} is factual in \textit{someone didn't forget to VP}, but not in \textit{someone forgot to VP}.

CB contains only \textit{VERB that S} frames. RP contains both \textit{VERB that S} and \textit{VERB to VP} frames.
MegaVeridicality exhibits nine frames, consisting of four argument structures
and manipulations of active/passive voice%
 and eventive/stative embedded VP:
\textit{VERB that S,
was VERBed that S,
VERB for NP to VP,
VERB NP to VP-eventive,
VERB NP to VP-stative,
NP was VERBed to VP-eventive,
NP was VERBed to VP-stative,
VERB to VP-eventive,
VERB to VP-stative}.

\paragraph{Annotation scales}
The original FactBank and MEANTIME annotations are categorical values.
We use \citet{stanovsky}'s unified representations for FactBank and MEANTIME,
which contain labels in the $[-3,3]$ range derived from to the
original categorical values in a rule-based manner.
The original annotations of MegaVeridicality contain three categorical values
\textit{yes/maybe/no}, which we mapped to $3$/$0$/$-3$ respectively. We then take the mean of the
annotations for each item.
The original annotations in RP are integers in $[-2, 2]$. We multiply each RP
annotation by 1.5 to get labels in the same range as in the other datasets.
The mean of the converted annotations is taken as the gold label for each item.

\section{Linguistic approaches to factuality}
Most work in NLP on event factuality has taken a lexicalist approach, tracing back factuality to fixed properties of lexical items. Under such an approach, properties of the lexical patterns present in the sentence determine the factuality of the event, without taking into account contextual factors. We will refer to the inference calculated from lexical patterns only as \textit{expected inference}. For instance, in \Next, the expected inference for the event \textit{had} embedded under
\textit{believe} is neutral. Indeed, since both true and false things can be
believed, one should not infer from \textit{A believes that S} that \textit{S}
is true (in other words, \textit{believe} has as \texttt{o/o} signature), making
\textit{believe} a so-called ``non-factive'' verb by opposition to ``factive'' verbs (such as \textit{know} or \textit{realize}, which generally entail the truth of their complements both in positive polarity sentences \ref{ex:pos} and in entailment-canceling environments \ref{ex:neg}, \citet{kiparsky1970}). However, lexical theories neglect the pragmatic enrichment that is pervasive in human communication and fall short in predicting the
correct inference in \Next, where people judged the content of the
complement to be true (as indicated by the annotation score of $2.38$).

{\small
\ex.\label{ex:annabel} Annabel \env{could} hardly \cev{believe} that she \event{had}{2.38} a daughter about to go to university.

}

In FactBank, \citet{sauri2009factbank} took a lexicalist approach, seeking to capture only the effect of lexical meaning and knowledge local to
the annotated sentence: annotators were linguistically trained and instructed to avoid using knowledge from the world or from the surrounding context of the sentence. However, it has been shown that  such annotations do
not always align with judgments from linguistically-naive annotators.
\citet{demarneffe2012} and \citet{lee2015event} re-annotated part of
FactBank with crowdworkers who were given minimal guidelines. They found that events embedded under report verbs (e.g., \textit{say}), annotated as neutral
in FactBank (since, similarly to \textit{believe}, one can report both true and false things), are often annotated as factual by crowdworkers. \citet{ross-pavlick-2019-well} showed that their annotations also exhibit such a \textit{veridicality bias}: events are often perceived as factual/nonfactual,
even when the expected inference specified by the signature is neutral.
The reason behind this misalignment is commonly attributed to pragmatics:
crowdworkers use various contextual features to perform pragmatic reasoning that overrides the
expected inference defined by lexical semantics. There has been theoretical
linguistics work arguing that factuality is indeed tied to the discourse structure and
not simply lexically controlled (i.a., \citealt{Simons10}).

\begin{figure*}
\resizebox{\textwidth}{!}{
  \centering
\begin{minipage}{\textwidth}
   \includegraphics[width=\textwidth]{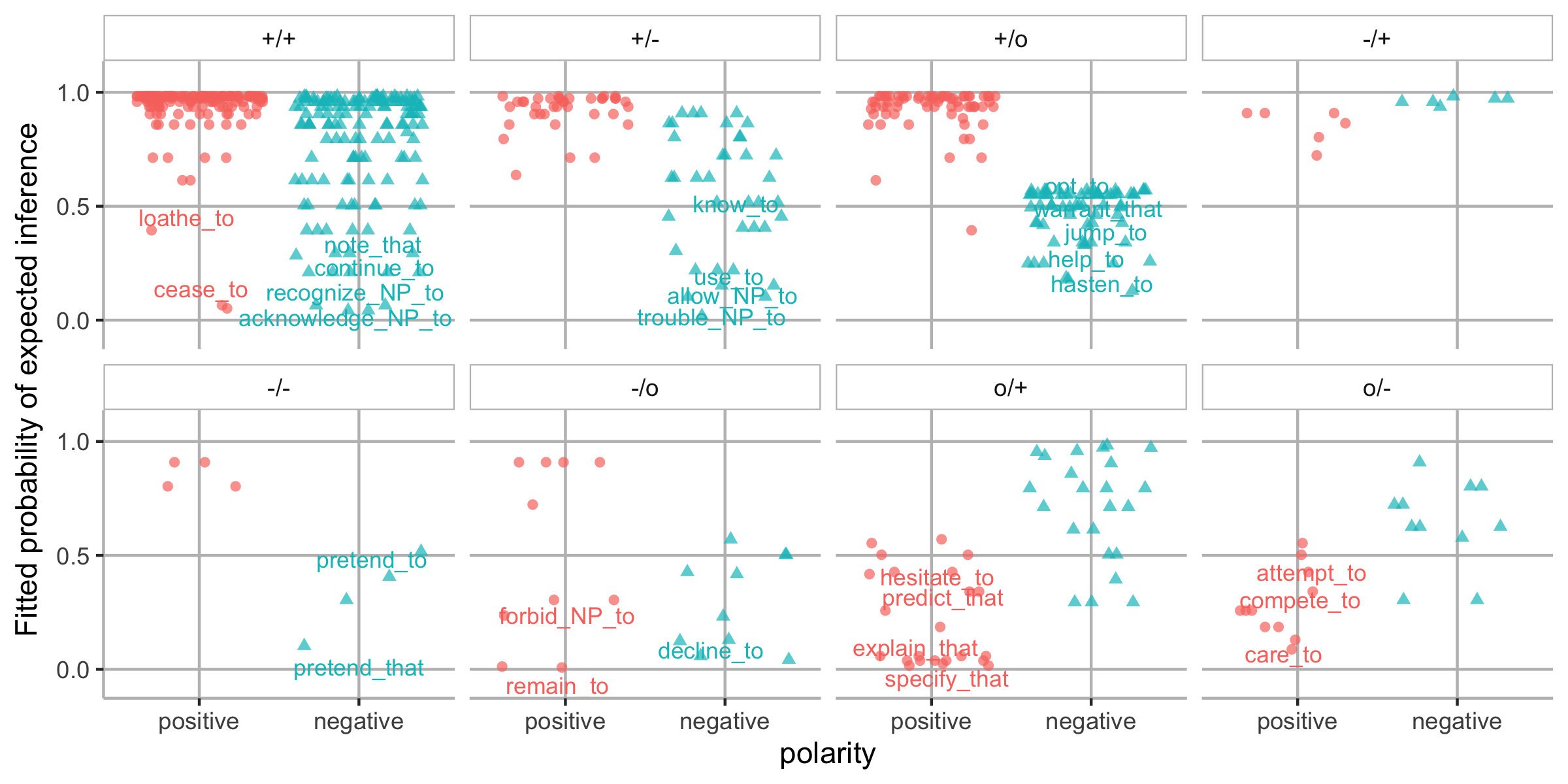}
   \caption{Fitted probabilities of true expected inference category predicted
     by the label of each item given by the ordered logistic regression model,
     organized by the signature and polarity.
     Some examples of verb-frame with mean probability less than 0.5 are labeled.}
  \label{fig:mv_ordinal_regression}
\end{minipage}
}
\end{figure*}

Further, our analysis of MegaVeridicality shows that
there is also some misalignment between the inference predicted by lexical semantics and the human annotations, even in cases without pragmatic factors. Recall that MegaVeridicality contains semantically bleached sentences where the only semantically loaded word is the embedding verb. We used ordered logistic regression to predict the expected inference category (\texttt{+}, \texttt{o}, \texttt{-}) specified by the embedding verb signatures defined in \citet{karttunen2012} from the mean human annotations.\footnote{The analysis is done on items with verb-frame combinations (and their
  passive counterparts) for which \citet{karttunen2012} gives a signature (i.e., 618 items from MegaVeridicality).}
The coefficient for mean human annotations is 1.488 (with 0.097 standard error): thus, overall, the expected inference aligns with the
annotations.\footnote{The threshold for \texttt{-|o} is -2.165 with SE 0.188. The
threshold for \texttt{o|+} is 0.429 with SE 0.144.}
However, there are cases where they diverge.
Figure~\ref{fig:mv_ordinal_regression} shows the fitted probability of the true
expected inference category for each item, organized by the signatures and polarity.
If the expected inference was always aligning with the human judgments, the fitted
probabilities would be close to 1 for all points.
However, many points have low fitted probabilities, especially when the
expected inference is \texttt{o} (e.g., negative polarity of \texttt{+/o} and \texttt{-/o}, positive polarity of \texttt{o/+} and
\texttt{o/-}), showing that there is veridicality bias in MegaVeridicality,
similar to RP.
Table~\ref{tab:examples_verb_mismatch_signature} gives concrete examples from MegaVeridicality and RP, for which the annotations often
differ from the verb signatures: events under \textit{not refuse to} are
systematically annotated as factual, instead of the expected neutral.
The RP examples contain minimal content information (but the mismatch in these examples may involve pragmatic reasoning).

\begin{table}[t]
  \centering
\resizebox{\linewidth}{!}{%
  \small
  \begin{tabular}[h]{p{10cm}}
\toprule
\textbf{not continue to} signature: \texttt{+/+}, expected: \texttt{+}, observed: \texttt{-}\\
    A particular person didn't continue to \event{do}{-0.33} a particular thing. \\
    A particular person didn't continue to \event{have}{-1.5} a particular thing. \\
    They did not continue to \event{sit}{-3} in silence.	\\
    He did not continue to \event{talk}{-3} about fish.	\\
\toprule
\textbf{not pretend to} signature: \texttt{-/-}, expected: \texttt{-}, observed: closer to \texttt{o}\\
    Someone didn't pretend to \event{have}{-1.2} a particular thing. \\
    He did not pretend to \event{aim}{-0.5} at the girls.	\\
    \toprule
  \textbf{\{add/warn\} that} signature: \texttt{o/+}, expected: \texttt{o}, observed: \texttt{+} \\
  Someone \cev{added} that a particular thing \event{happened}{2.1}. \\
   Linda Degutis \cev{added} that interventions \event{have}{2.5} to be monitored.	\\
  Someone \cev{warned} that a particular thing \event{happened}{2.1}. \\
  It \cev{warns} that \event{Mayor Giuliani 's proposed pay freeze could destroy the NYPD 's new esprit de corps}{2.5}. \\
  \toprule
\textbf{not \{decline/refuse\} to} signature: \texttt{-/o}, expected: \texttt{o}, observed: \texttt{+}\\
A particular person \env{didn't} \cev{decline} to \event{do}{1.5} a particular thing. \\
We do \env{not} \cev{decline} to \event{sanction}{2.5} such a result. \\
A particular person \env{didn't} \cev{refuse} to \event{do}{2.1} a particular thing. \\
The commission did \env{not} \cev{refuse} to \event{interpret}{2.0} it. \\
  \bottomrule

\end{tabular}
}
\caption{Items with verbs that often behave differently from the
  signatures.
  The semantically-bleached sentences are from MegaVeridicality, the others from RP.
  Gold labels are superscripted.}
\label{tab:examples_verb_mismatch_signature}
\end{table}

In any case, given that neural networks are function approximators, we hypothesize that
BERT can learn these surface-level lexical patterns in the training data.
But items where pragmatic reasoning overrides the lexical patterns would probably
be challenging for the model.

\section{Model and experiment setup}
\label{sec:experiment}
To analyze what BERT can learn, we use the seven factuality datasets in
Table~\ref{tab:examples}.

\paragraph{Data preprocessing}
  \begin{table}[t!]
    \centering
    \small
    \resizebox{0.8\linewidth}{!}{
  \begin{tabular}[t!]{lrrr}
    \toprule
                     & train  & dev   & test  \\\midrule
    MegaVeridicality & 2,200   & 626   & 2,200 \\
    CommitmentBank   & 250    & 56    & 250   \\
    RP               & 1,100  & 308   & 1,100 \\
    \midrule
    FactBank         & 6,636  & 2,462 & 663   \\
    MEANTIME         & 1,012  & 195   & 188   \\
    UW               & 9,422  & 3,358 & 864   \\
    UDS-IH2          & 22,108 & 2,642 & 2,539 \\
    \bottomrule
  \end{tabular}
  }
  \captionof{table}{
    Number of events in each dataset split.
  }
  \label{tab:size}
\end{table}

The annotations of CB and RP have been collected by asking annotators to rate the
factuality of the content of the complement, which may contain other polarity and modality
operators, whereas in FactBank annotators rated the
factuality of the normalized complement, without polarity and modality operators.
For example, the complement
\textit{anything should be done in the short term} contains the modal operator
\textit{should}, while the normalized complement would be \textit{anything is
  done in the short term}.
In MEANTIME, UW and UDS-IH2, annotators rated the factuality of the event represented by a
word in the original sentence, which has the effect of removing such operators.
Therefore, to ensure a uniform interpretation of annotations between datasets, we semi-automatically identified items in CB and RP where the complement is
not normalized,\footnote{We automatically identified whether the complement
  contains a \texttt{neg} dependency relation, modal operators (\textit{should, could, can, must, perhaps, might,
    maybe, may, shall, have to, would}), or adverbs,
and manually verified the output.} for which we take the whole embedded clause to be the span for factuality prediction.
Otherwise, we take the root of the embedded clause as the span.

We also excluded 236 items in RP where the event for which annotations were gathered cannot be represented by a
single span from the sentence. For example, for \textit{The Post Office is forbidden from ever attempting to close any office},
annotators were asked to rate the factuality of \textit{the Post Office is forbidden from ever closing any
  office}. Simply taking the span \textit{close any office} corresponds to
the event of \textit{the Post Office close any office}, but not to the event for
which annotations are collected.

\paragraph{Excluding data with low agreement annotation}
There are items in RP and CB which exhibit bimodal annotations. For instance, the sentence
in RP
\textit{White ethnics have ceased to be the dominant force in urban life}
received 3 annotation scores: $-3$/nonfactual, $1.5$/between neutral and factual, and $3$/factual.
By taking the mean of such bimodal annotations, we end up with a label of $0.5$/neutral,
which is not representative of the judgments in the individual annotations.
For RP (where each item received three annotations), we excluded 250 items where at least two annotations have different signs.
For CB (where each item received at least 8 annotations), we follow \citet{jiang-de-marneffe-2019-know} by binning the responses into $[-3, -1]$,
$[0]$, $[1,3]$ and discarding items if less than 80\% of the annotations fall in the same bin.

\paragraph{Data splits}
We used the standard train/dev/test split for FactBank, MEANTIME, UW, and UDS-IH2.
As indicated above, we only use the high agreement subset of CB with 556 items,
with splits from \citet{jiang-de-marneffe-2019-evaluating}.
We randomly split MegaVeridicality and RP with stratified sampling to keep the distributions of the clause-embedding verbs similar in each split. Table~\ref{tab:size} gives the number of items in each split.

\paragraph{Model architecture}
The task is to predict a scalar value in $[-3, 3]$ for each event described by
a span in the input sentence.
A sentence is fed into BERT and the final-layer representations for the
event span are extracted.
Since the spans have variable lengths, the SelfAttentiveSpanExtractor \citep{gardner-etal-2018-allennlp} is used to weightedly combine the representations of multiple tokens and create a single vector for the original event span.
The extracted span vectors are fed into a two-layer feed-forward network with tanh activation function to predict a single scalar value. Our architecture is similar to \citet{rudinger}'s linear-biLSTM model, except that the input is encoded with BERT instead of bidirectional LSTM, and a span extractor is used.
The model is trained with the smooth L1 loss.\footnote{The code and data are available at \url{https://github.com/njjiang/factuality_bert}. The code is based on the toolkit \texttt{jiant v1} \citep{wang2019jiant}.}

\paragraph{Evaluation metrics}
Following previous work, we report mean absolute error (MAE), measuring absolute fit, and Pearson's $r$ correlation, measuring how well models capture variability in the data. $r$ is considered more informative since some datasets (MEANTIME in particular) are biased towards $+3$.

\paragraph{Model training}
For all experiments, we fine-tuned BERT using the \texttt{bert\_large\_cased} model. Each model is fine-tuned with at most 20 epochs, with a learning rate of $1e-5$.
Early stopping is used: training stops if the difference between Pearson's $r$ and MAE
does not increase for more than 5 epochs.
Most training runs last more than 10 epochs.
The checkpoint with the highest difference between Pearson's $r$ and MAE on the dev set is used for testing. We explored several training data combinations:\\
\textbf{-Single:} Train with each dataset individually;\\
\textbf{-Shared:} Treat all datasets as one;\\
\textbf{-Multi:} Datasets share the same BERT parameters while each
  has its own classifier parameters.

The Single and Shared setups may be combined with first fine-tuning BERT on MultiNLI, denoted by the superscript $^M$. We tested on the test set of the respective datasets.

We also tested whether BERT improves on previous models on its ability to generalize to embedded events.
The models in \citet{rudinger} were trained on FactBank, MEANTIME, UW, and UDS-IH2 with shared encoder parameters and separate classifier parameters, and an ensemble of the four classifiers.
To make a fair comparison, we followed \citeauthor{rudinger}'s setup by training BERT on FactBank, MEANTIME, UW, and UDS-IH2 with one single set of parameters\footnote{Unlike the Hybrid model of \citet{rudinger}, there is no separate classifier parameters for each dataset.} and tested on MegaVeridicality and CommitmentBank.\footnote{For both datasets, examples from all splits are used, following previous work.}

\begin{table*}[t!]
\resizebox{\linewidth}{!}{
\begin{tabular}{l|lllllp{1.2cm}|lllllp{1.2cm}}
\toprule

  & \multicolumn{5}{l}{\textbf{R}}  & Previous & \multicolumn{5}{l}{\textbf{MAE}} & Previous \\
 & Shared     & Shared\M & Single & Single\M & Multi      &  SotA & Shared    & Shared\M & Single & Single\M & Multi      & SotA\\
  \midrule
CB               & 0.865                      & 0.869                      & 0.831                      & 0.878    & \cellcolor[HTML]{B0C4DE}0.89$^\dagger$ &                            & 0.713  & 0.722                      & 0.777                      & 0.648    & \cellcolor[HTML]{B0C4DE}0.617$^\dagger$ &                            \\
RP               & 0.806                      & 0.813                      & 0.867                      & 0.867    & \cellcolor[HTML]{B0C4DE}0.87$^\dagger$ &                            & 0.733  & 0.714                      & 0.621                      & 0.619    & \cellcolor[HTML]{B0C4DE}0.608$^\dagger$ &                            \\
MegaVeridicality & \cellcolor[HTML]{B0C4DE}0.876$^\dagger$ & 0.873                      & 0.857                      & 0.863    & 0.857                     &                            & 0.508  & \cellcolor[HTML]{B0C4DE}0.501$^\dagger$ & 0.531                      & 0.523    & 0.533                      &                            \\
FactBank         & 0.836                      & 0.845                      & \cellcolor[HTML]{B0C4DE}0.914$^\dagger$ & 0.901    & 0.903                     & 0.903                         & 0.42   & 0.417                      & \cellcolor[HTML]{B0C4DE}0.228$^\dagger$ & 0.241    & 0.236                      & 0.31                          \\
MEANTIME         & 0.557                      &0.572$^\dagger$ & 0.503                      & 0.513    & 0.491                     & \cellcolor[HTML]{B0C4DE}0.702 & 0.333  & 0.338                      & 0.355                      & 0.345    & 0.319$^\dagger$ & \cellcolor[HTML]{B0C4DE}0.204 \\
UW               & 0.776                      & 0.787                      & \cellcolor[HTML]{B0C4DE}0.868$^\dagger$ & 0.868    & 0.865                     & 0.83                          & 0.532  & 0.523                      & \cellcolor[HTML]{B0C4DE}0.349$^\dagger$ & 0.351    & 0.351                      & 0.42                          \\
  UDS-IH2          & 0.845                      & 0.843                      & 0.855$^\dagger$ & 0.854    & 0.853                     & \cellcolor[HTML]{B0C4DE}0.909 & 0.794  & 0.804                      & 0.76$^\dagger$  & 0.763    & 0.766                      & \cellcolor[HTML]{B0C4DE}0.726\\
  \bottomrule
\end{tabular}
}
\caption{Performance on the test sets under different BERT training setups.
  The best score obtained by our models for each dataset under each metric is
  marked by $\dagger$.
  The overall best scores are highlighted.
  Each score is the average from three runs with different random initialization.
  The previous state-of-the-art results are given when available. All come from \citet{pouran-ben-veyseh-etal-2019-graph}, except the MAE score on UW which comes from \citet{stanovsky}.
}
\label{tab:scores}
\end{table*}

\begin{table}[t!]
  \centering
  \small
  \resizebox{\linewidth}{!}{
  \begin{tabular}[t!]{lllll}
    \toprule
 & \multicolumn{2}{l}{MegaVeridicality} & \multicolumn{2}{l}{CB} \\
               & $r$  & MAE  & $r$  & MAE  \\
    \midrule
    BERT                   & 0.60 & 1.09 & 0.59 & 1.40  \\
    \midrule
    \citeauthor{stanovsky} & -    & -    & 0.50  & 2.04  \\
    \citeauthor{rudinger}  & 0.64 & -    & 0.33 & 1.87  \\
    \bottomrule
  \end{tabular}
  }
  \captionof{table}{Performance on MegaVeridicality and CommitmentBank across all splits
    of the previous model  (\citealt{stanovsky}
    and \citealt{rudinger}) and BERT trained on the concatenation of FactBank, MEANTIME,
    UW, UDS-IH2 using one set of parameters.
    \citet{white-etal-2018-lexicosyntactic} did not report MAE results for MegaVeridicality.}
  \label{tab:cb556}
\end{table}

\section{Results}
Table~\ref{tab:scores} shows
performance on the various test sets with the different training schemes. These
models perform well and obtain the new state-of-the-art results on FactBank and UW, and
comparable performance to the previous models on the other
datasets (except for MEANTIME\footnote{The difference in performance for MEANTIME might come from a difference in splitting: \citet{pouran-ben-veyseh-etal-2019-graph}'s test set has a different size. Some of the gold labels in MEANTIME also seem wrong.}).
Comparing Shared vs.\ Shared\M and Single vs.\ Single\M,
we see that transferring with MNLI helps all datasets on at least one metric,
except for UDS-IH2 where MNLI-transfer hurts performance.
The Multi and Single models obtain the best performance on almost all datasets
other than MegaVeridicality and MEANTIME.
The success of these models confirms the findings of \citet{rudinger} that having dataset-specific parameters is necessary for optimal performance.
Although this is expected, since each dataset has its own specific features,
the resulting model captures data-specific quirks rather than generalizations about event factuality. This is problematic if one wants to deploy the system in downstream applications,
since which dataset the input sentence will be more similar to is unknown a priori.

However, looking at whether BERT improves on the previous
state-of-the-art results for its ability to generalize to the linguistic
constructions without in-domain supervision, the results are less promising.
Table~\ref{tab:cb556} shows performance of BERT trained on four factuality
datasets
and tested on MegaVeridicality and CB across all splits, and the Rule-based and Hybrid
models' performance reported in \citet{jiang-de-marneffe-2019-know} and
\citet{white-etal-2018-lexicosyntactic}.
BERT improves on the other systems by only a small margin for CB, and obtains no
improvement for MegaVeridicality. Despite having a magnitude more parameters and pretraining, BERT does not generalize to the embedded events present in MegaVeridicality and CB.
This shows that we are not achieving robust natural language understanding,
unlike what the near-human performance on various NLU benchmarks suggests.

Finally, although RoBERTa \cite{roberta} has exhibited improvements over BERT on many different tasks, we found that, in this case, using pretrained RoBERTa instead of BERT does not yield much improvement.
The predictions of the two models are highly correlated, with 0.95 correlation over all datasets' predictions.

\section{Quantitative analysis: Expected inference}
\label{sec:error_analylsis}

Here, we evaluate our hypothesis that BERT can learn subtle lexical patterns,
regardless of whether they align with lexical semantics theories,
but struggles when pragmatic reasoning overrides the lexical patterns.
To do so, we present results from a quantitative analysis using the notion
of expected inference.
To facilitate meaningful analysis, we generated two random train/dev/test
splits of the same sizes as in Table~\ref{tab:size} (besides the standard split) for MegaVeridicality, CB, and RP.
All items are present at least once in the test sets.
We trained the Multi model using three different random initializations with
each split.\footnote{There is no model performing radically better than the others. The Multi model achieves better results than the Single one on CB and is getting comparable performance to the Single model on the other datasets.} We use the mean predictions of each item across all initializations and all splits (unless
stated otherwise).

\subsection{Method}
\label{sec:quant_method}
As described above, the expected inference of an item is the factuality label
predicted by lexical patterns only.
We hypothesize that BERT does well on items where the gold labels match the expected inference, and fails on those that do not.

\paragraph{How to get the best expected inference?}
To identify the expected inference, the approach varies by dataset.
For the datasets focusing on embedded events (MegaVeridicality, CB, and RP),
we take, as expected inference label, the mean labels of training items with the same combination of features as the test item.
Theoretically, the signatures should capture the expected inference. However, as shown earlier, the signatures do not always align with the observed annotations, and not all verbs have signatures defined.
The mean labels of training items with the same features captures what the common patterns in the data are and what the model is exposed to.
In MegaVeridicality and RP,
the features are clause-embedding verb, polarity and frames. In CB, they are verb and entailment-canceling
environment.\footnote{
  The goal is to take items with the most matching features.
If there are no training items with the exact same combination of features,
we take items with the next best match, going down the list if the previous features are not available:\\
- MegaVeridicality and RP: verb-polarity, verb, polarity. \\
- CB: verb, environment.
}

For FactBank, UW, and MEANTIME, the approach above does not apply because these
datasets contain matrix-clause and embedded events. %
We take the predictions from \citeauthor{stanovsky}'s Rule-based model\footnote{\url{https://github.com/gabrielStanovsky/unified-factuality/tree/master/data/predictions_on_test}} as the expected inference,
since the Rule-based model uses lexical rules including the signatures.
We omitted UDS-IH2 from this analysis because there are no existing
predictions by the Rule-based model on UDS-IH2 available.

\subsection{Results}
We fitted a linear mixed effect model using the absolute error between the
expected inference and the label to predict the absolute error of the model
predictions, with random intercepts and slopes for each dataset.
Results are shown in Table~\ref{tab:lm_inf}.
We see that the slopes are all positive, suggesting that
the error of the expected inference to the label is positively correlated with
the error of the model,
as we hypothesized.

\begin{table}[t]
\centering
\resizebox{\linewidth}{!}{
\begin{tabular}{lrrrr}
\toprule
Dataset  & $\alpha$ & $SE(\alpha)$ & $\beta$ & $SE(\beta)$ \\
\midrule
FactBank         & -0.039 & 0.018 & 0.073 & 0.015 \\
MEANTIME         & -0.058 & 0.033 & 0.181 & 0.024 \\
UW               & 0.004  & 0.016 & 0.261 & 0.016 \\
MegaVeridicality & 0.134  & 0.008 & 0.142 & 0.006 \\
CB               & 0.099  & 0.020 & 0.265 & 0.016 \\
RP               & 0.059  & 0.011 & 0.468 & 0.012 \\
\bottomrule
\end{tabular}}
\caption{Estimated random intercepts ($\alpha$) and slopes ($\beta$) for each dataset and their standard errors. The fixed intercept is 0.228 with standard error 0.033.
}
\label{tab:lm_inf}
\end{table}

The slope for FactBank is much smaller than the slopes for the other datasets,
meaning that for FactBank, the error of the expected inference does not predict the model's errors as much as in the other datasets.
This is due to the fact that the errors in FactBank consist of items for which the lexicalist and crowdsourced annotations may differ. The model, which has been trained
on crowdsourced datasets, makes predictions that are more in line with the crowdsourced annotations but are errors compared to the lexicalist labels.
For example, 44\% of the errors are reported events (e.g., \textit{X said that \ldots}) annotated
as neutral in FactBank (given that both true or false things can be reported) but predicted as factual. Such reported events have been found to be annotated as factual by crowdworkers (\citealt{demarneffe2012}, \citealt{lee2015event}).
On the other hand, the expected inference (from the Rule-based model) also follows a lexicalist
approach. Therefore labels align well with the expected inference, but the predictions do so poorly.

\section{Qualitative analysis}
The quantitative analysis shows that the model predictions are driven by surface-level features. Not surprisingly, when a gold label of an item diverges from the label of items with similar surface patterns, the model does not do well. Here, we unpack which surface features are associated with labels, and examine the edge cases in which surface features diverge from the observed labels.
We focus on the CB, RP, and MegaVeridicality datasets since they focus on embedded events well studied in the literature.

\begin{figure*}
  \centering
    \includegraphics[width=\textwidth]{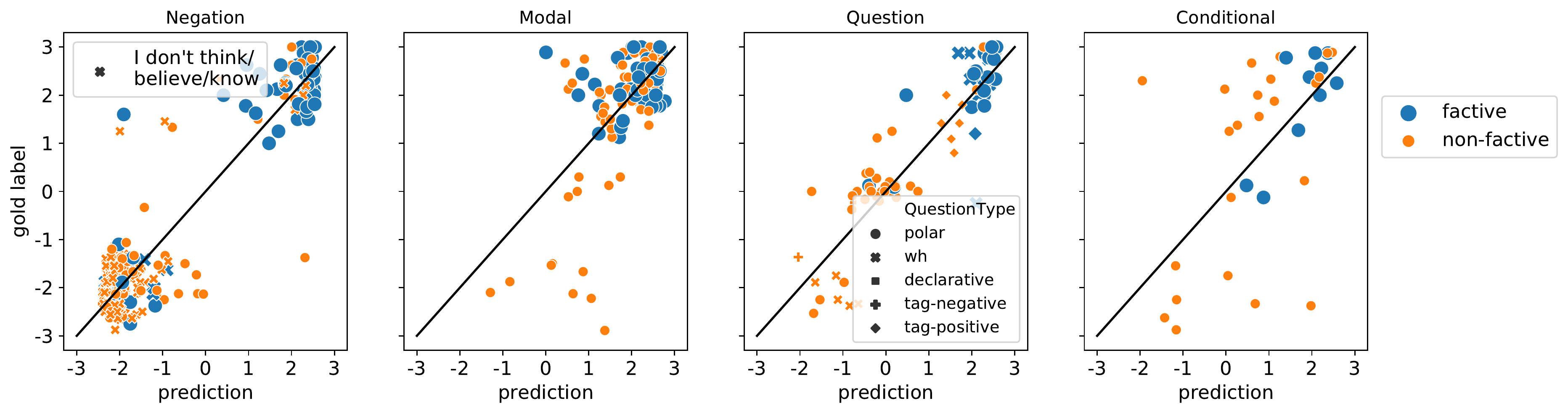}
    \captionof{figure}{Multi model's predictions compared
      to gold labels for all CB items present in all splits, by
      entailment-canceling environment. Diagonal line shows perfect prediction.
    }
  \label{fig:cb_scatter}
\end{figure*}

\subsection{CB}
Figure~\ref{fig:cb_scatter} shows the scatterplot of the Multi model's prediction
vs.\ gold labels on CB, divided by each entailment-canceling environment.
As pointed out by \citet{jiang-de-marneffe-2019-evaluating},
the interplay between the entailment-canceling environment and the clause-embedding verb is often the deciding factor for the factuality of the complement in CB.
Items with factive embedding verbs tend indeed to be judged as factual (most blue points in Figure~\ref{fig:cb_scatter} are at the top of the panels).
``Neg-raising'' items contain negation
in the matrix clause (\textit{not \{think/believe/know\} $\phi$}) but are
interpreted as negating the content of the complement clause (\textit{\{think/believe/know\} not $\phi$}). Almost all items involving a construction indicative of ``Neg-raising'' \textit{I don’t think/believe/know $\phi$} have nonfactual labels (see \texttimes\
 in first panel of Figure~\ref{fig:cb_scatter}).
Items in modal environment are judged as factual (second panel where most points are at the top).

In addition to the environment and the verb, there are more fine-grained surface patterns predictive of human annotations.
Polar question items with nonfactive verbs often have near-0 factuality labels (third panel, orange circles clustered in the middle). In tag-question items, the label of the embedded event often matches the matrix clause polarity, such as \ref{ex:tagq} with a matrix clause of positive polarity and a factual embedded event.

{\small
\ex.\label{ex:tagq} [\ldots] I \cev{think} it \eventp{went}{1.09}{1.52} to Lockheed, \cev{didn't it}?

}

Following these statistical regularities, the model obtains good results by
correctly predicting the majority cases.
However, it is less successful on cases where the surface features do not lead
to the usual label, and pragmatic reasoning is required. The model predicts most of the neg-raising items correctly, which make up 58\% of the data under negation.
But the neg-raising pattern leads the model to predict negative
values even when the labels are positive, as in \Next.\footnote{We use the notation \eventp{event span}{label}{prediction} throughout the rest of the paper.}

{\small
  \ex.
    [\ldots] And I think society for such a long time said, well, you know, you're married, now
   you need to have your family and I \env{don't}
   \cev{think} it's
   \eventp{been}{1.25}{-1.99} until recently that they had decided that two people was a
   family.

   }
It also wrongly predicts negative values for items where the context
contains a neg-raising-like substring (\textit{don't think/believe}), even when
the targeted event is embedded under another environment: question for \Next,
antecedent of conditional for \NNext.

{\small
\ex. B: All right, well. A: Um, short term, I don't think anything's going to be
done about it or probably should be done about it. B: Right. Uh, \env{are you
  saying} you \uwave{\env{don't} \cev{think}} \eventp{anything should be done in
  the short term}{0}{-1.73}?

\ex.   [\ldots] %
I \uwave{do not believe} I am being unduly boastful \env{if}
I \cev{say} that very few ever \eventp{needed}{2.3}{-1.94} amendment.

}

\begin{figure*}[t]
  \centering
    \includegraphics[width=\textwidth]{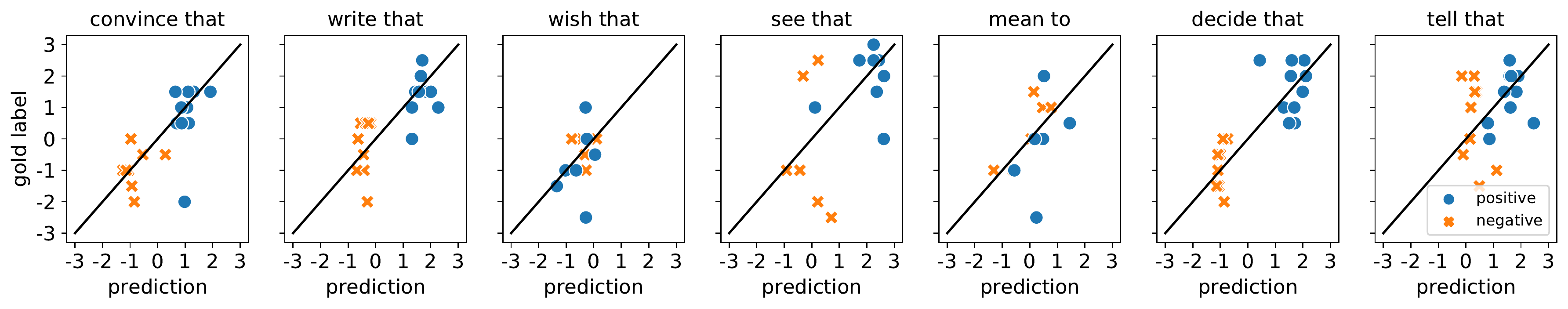}
    \captionof{figure}{Multi model's predictions compared
      to gold labels for certain verbs and frames in RP.
Diagonal line shows perfect prediction.
    }
  \label{fig:rp_scatter}
\end{figure*}

\subsection{RP}

The surface features impacting the annotations in RP are the clause-embedding verb, its syntactic frame, and polarity.
Figure~\ref{fig:rp_scatter} shows the scatterplot of label vs.\ prediction for
items with certain verbs and frames, for which we will show concrete
examples later. The errors (circled points) in each panel are
often far away from the other points of the same polarity on the $y$-axis,
confirming the findings above that the model fails on items that diverge from items with similar surface patterns.
Generally, points are more widespread along the $y$-axis than the $x$-axis, meaning that
the model makes similar predictions for items which share the same features, but it cannot account for variability among such items.
Indeed, the mean variance of the predictions for items of each verb, frame, and polarity is 0.19, while the mean variance of the gold labels for these items is 0.64.

Compare \ref{ex:convince_alarm} and \ref{ex:convince_soho}: they consist of the same
verb \textit{convince} with positive polarity and they have similar predictions, but very different gold labels.
Most of the \textit{convince} items of positive polarity are between neutral and
factual (between $0$ and $1.5$), such as \ref{ex:convince_alarm}. The model learned that from the training data: all \textit{convince} items of positive polarity have similar predictions ranging from
$0.7$ to $1.9$, with mean $1.05$ (also shown in the first panel of Figure~\ref{fig:rp_scatter}).
However, \ref{ex:convince_soho} has a negative label of $-2$ unlike the other
\textit{convince} items, because the following context \textit{I was mistaken}
clearly states that the speaker's belief is false, and therefore the event \textit{they would fetch up at the house in Soho} is not factual. Yet the model fails to take this into account.

{\small
\ex. \label{ex:convince_alarm} I was \cev{convinced} that \eventp{the alarm was given when Mrs. Cavendish was in the room}{1.5}{1.13}.

\ex.\label{ex:convince_soho} I was \cev{convinced} that \eventp{they would fetch up at the house in Soho}{-2}{0.98}, but it appears I was mistaken.

}

\subsection{MegaVeridicality}
As shown in the expected inference analysis, MegaVeridicality exhibits the same error pattern as CB and RP (failing on items where gold labels differ from the ones of items sharing similar surface features).
Unlike CB and RP, MegaVeridicality is designed to rule out the effect of pragmatic reasoning. Thus the errors for MegaVeridicality cannot be due to pragmatics. Where are those stemming from?
It is known that some verbs behave very differently in different frames. However, the model was not exposed to the same combination of verb and frame during training and testing, which leads to errors.
For example, \textit{mislead}\footnote{Other verbs with the same behavior and similar
  meaning include \textit{dupe, deceive, fool}.} in the \textit{VERBed NP to VP}
frame in positive polarity, as in \ref{ex:mislead_NP_to_do},
and its passive counterpart \ref{ex:was_misled_to_do},
suggests that the embedded event is factual (\textit{someone did something}),
while in other frame/polarity, the event is nonfactual, as in \ref{ex:was_misled_that} and \ref{ex:wasnt_misled_to_do}.
The model, following the patterns of \textit{mislead} in other contexts, fails on \ref{ex:mislead_NP_to_do} and \ref{ex:was_misled_to_do} because the
training set did not contain
instances with \textit{mislead} in a factual context.

{\small
\ex.\label{ex:mislead_NP_to_do} Someone \cev{misled a particular person to} \eventp{do}{2.7}{-1.6} a particular thing.

\ex.\label{ex:was_misled_to_do} A particular person \cev{was misled to} \eventp{do}{2.7}{-1.21} a particular thing.

\ex.\label{ex:was_misled_that} Someone \cev{was misled that} a particular thing \eventp{happened}{-1.5}{-2.87}.

\ex.\label{ex:wasnt_misled_to_do} Someone \cev{was}\env{n't} \cev{misled to} \eventp{do}{-0.3}{-0.6} a particular thing.

}

This shows that the model's ability to reason is still limited to pattern
matching: it fails to induce how verb meaning interacts with syntactic frames that are unseen during training.
If we augment MegaVeridicality with more items of verbs in these contexts (currently there is one example of each verb under either polarity in most frames) %
and add them to the training set, BERT would probably learn these behaviors. %

Moreover, the model here exhibits a different pattern from \citet{white-etal-2018-lexicosyntactic}, who found that their model cannot capture inferences whose polarity mismatches the matrix clause polarity, as their model
fails on items with verbs that suggest nonfactuality of their complement such as \textit{fake, misinform} under positive polarity.
As shown in the expected inference analysis in section \ref{sec:error_analylsis}, our model is successful at these items, since it has memorized the lexical pattern in the training data.

\subsection{Error categorization}
\label{sec:error_categorization}
In this section, we study the kinds of reasoning that is needed to draw the
correct inference in items that the system does not handle correctly.
For the top 10\% of the items sorted by absolute error in CB and RP,
two linguistically trained annotators annotated which factors lead to the
observed factuality inferences, according to factors put forth in the literature, as described below.\footnote{This is not an exhaustive list of reasoning types present in the data, and having one of these properties is not sufficient for the model to fail.}

\paragraph{Prior probability of the event}
Whether the event described is likely to be true is
known to influence human judgments of event factuality
\citep{10.1093/jos/ffy007,demarneffe2018}. %
Events that are more likely to be factual a priori are often considered as factual
even when they are embedded, as in \ref{ex:carriages}. Conversely, events that are unlikely a priori are rated as nonfactual when embedded, as in \ref{ex:mexico}.

{\small
\ex.\label{ex:carriages} %
[\ldots] He took the scuffed leather
document case off the seat beside him and banged the door shut with the violence
of someone who had \env{not} \cev{learned} that \eventp{car doors do not need the same sort
  of treatment as those of railway carriages}{2.63}{0.96}

\ex.\label{ex:mexico} In a column lampooning Pat Buchanan, Royko did \env{not} \cev{write} that Mexico
\eventp{was}{-3}{-0.3} a useless country that should be invaded and turned over to Club
Med.

}

\paragraph{Context suggests (non)factuality}
The context may directly describe or give indirect cues about the factuality of the content of the complement. In \ref{ex:wish_french}, the preceding context \textit{they're French} clearly indicates that the content of the complement is false.
The model predicts $-0.28$ (the mean label for training items with \textit{wish} under positive polarity is $-0.5$), suggesting that the model fails to take the preceding context into account.

{\small
\ex.\label{ex:wish_french} They're French, but \cev{wish} that they \eventp{were}{-2.5}{-0.28} mostly Caribbean.

}

The effect of context can be less explicit, but nonetheless there. In \ref{ex:believe_warnings}, the context \textit{which it's mainly just when it gets real, real
  hot} elaborates on the time of the warnings, carrying the
presupposition that the content of the complement \textit{they have warnings here} is true.
In \ref{ex:tarzan}, the preceding context \textit{Although Tarzan is now nominally in
  control}, with the marker \textit{although} and \textit{nominally} suggesting that
Tarzan is not actually in charge, makes the complement \textit{Kala the Ape-Mom is really in charge} more likely.

{\small
  \ex.\label{ex:believe_warnings}
  [...] %
B: Oh, gosh, I think I would hate to live in California, the smog there.
A: Uh-huh. B: I mean, I \env{can't} \cev{believe} they \eventp{have}{2.33}{0.327} warnings here, which it's mainly just when it gets real, real hot.

\ex. \label{ex:tarzan}
Although Tarzan is now nominally in control, one does \env{not} \cev{suspect} that Kala the Ape-Mom, the Empress Dowager of the Jungle, \eventp{is}{2.5}{-0.23} really in charge.

}

\paragraph{Discourse function}
When sentences are uttered in a discourse, there is a discourse goal or a
question under discussion (QUD) that the sentence is trying to address \cite{roberts:2012:information}.
According to \citet{10.1093/jos/ffy007}, the contents of embedded complements that do not address the
question under discussion are considered as more factual than those that do address the QUD.
Even for items that are sentences in isolation, as in RP, readers interpreting these sentences probably reconstruct a discourse and the implicit QUD that the sentences are trying to address.
For instance, \ref{ex:jon} contains the factive verb \textit{see}, but its complement is labeled as nonfactual ($-2$).

{\small
\ex. \label{ex:jon}Jon did \env{not} \cev{see} that they \eventp{were}{-2}{1.45} hard pressed.

}

Such label is compatible with a QUD asking what is the evidence that Jon has to whether they were hard pressed. The complement does not answer that QUD, but the sentence affirms that Jon lacks visual evidence to conclude that they were hard pressed.
In \ref{ex:vice_told}, the embedded event is annotated as factual although it is embedded under a report verb (\textit{tell}). However, the sentence in \ref{ex:vice_told} can be understood as providing a partial answer to the QUD
\textit{What was the vice president told?}. The content of the
complement does not address the QUD, and is therefore perceived as factual.

{\small
\ex. \label{ex:vice_told} The Vice President was \env{not} \cev{told} that the Air Force was \eventp{trying}{2}{-0.15} to protect the Secretary of State through a combat air patrol over Washington.

}

\paragraph{Tense/aspect}
The tense/aspect of the clause-embedding verb and/or the complement affects the
factuality of the content of the complement \cite{doi:10.1080/08351817109370248, demarneffe2018}. In \ref{ex:meant_warn}, the past perfect \textit{had meant} implies that the complement did not happen ($-2.5$), whereas in \ref{ex:means} in the present tense, the complement is interpreted as neutral ($0.5$).

{\small
\ex.\label{ex:meant_warn} She had \cev{meant} to \eventp{warn}{-2.5}{-0.24} Mr. Brown about Tuppence.

\ex.\label{ex:means} A bigger contribution \cev{means} to \eventp{support}{0.5}{1.45} candidate Y.

}

\paragraph{Subject authority/credibility}
The authority of the subject of the clause-embedding verb also affects factuality judgments (i.a., \citealt{schlenker2010local}, \citealt{demarneffe2012}).
The subjects of \Next, a legal document, and \NNext, the tenets of a religion, have the authority to require or demand. Therefore what the legal document requires is perceived as factual, and
what the tenets do not demand is perceived as nonfactual.

{\small
\ex. Section 605(b) \cev{requires} that the Chief Counsel \eventp{gets}{2.5}{0.53} the
statement.

\ex. The tenets of Jainism do \env{not} \cev{demand} that \eventp{everyone must be wearing
  shoes when they come into a holy place}{-2}{-0.28}.

}

On the other hand, the perceived lack of authority of the subject may suggest that the embedded event is not factual.
In \ref{ex:witness}, although \textit{remember} is a factive verb, the embedded event only receives a mean annotation of $1$, probably because the subject \textit{a witness} introduces a specific situational context questioning whether to consider someone's memories as facts.

{\small
\ex. \label{ex:witness}
A witness \cev{remembered} that there \eventp{were}{1}{2.74} four simultaneous decision making processes going on at once.

}

\paragraph{Subject-complement interaction for prospective events}
Some clause-embedding verbs, such as \textit{decide} and \textit{choose}, introduce so-called ``prospective
events'' which could take place in the future \citep{sauri:guideline}. The
likelihood that these events will actually take place depends on several
factors: the content of the complement itself, the embedding verb and the
subject of the verb. When the subject of the clause-embedding verb is the same
as the subject of the complement, the prospective events are often judged as
factual, as in \ref{ex:stone}. In \ref{ex:poirot}, the subjects of the main verb and the complement verb are different, and the complement is judged as neutral.

{\small
\ex.\label{ex:stone} He \cev{decided} that \eventp{he must leave no stone unturned}{2.5}{0.43}.

\ex.\label{ex:poirot} Poirot \cev{decided} that \eventp{Miss Howard must be kept in the dark}{0.5}{1.49}.

}

Even when subjects are the same, the nature of the prospective event itself also affects whether it is perceived as factual. Compare
\ref{ex:clinton} and \ref{ex:buried} both featuring the construction \textit{do not choose to}: \ref{ex:clinton} is judged as nonfactual whereas \ref{ex:buried} is neutral. This could be due to the difference in the extent to which
the subject entity has the ability to fulfill the chosen course of action
denoted by the embedded predicate.
In \ref{ex:clinton}, Hillary Clinton can be perceived to be
able to decide where to stay, and therefore when she does not choose to stay
somewhere, one infers that she indeed does not stay there.
On the other hand, the subject in \ref{ex:buried} is not able to fulfill the
chosen course of action (where to be buried), since he is presumably dead.

{\small
\ex. \label{ex:clinton} Hillary Clinton does \env{not} \cev{choose} to \eventp{stay}{-2.5}{-0.92} at Trump Tower.

\ex. \label{ex:buried} He did \env{not} \cev{choose} to be \eventp{buried}{0.5}{-0.75} there.

}

\paragraph{Lexical inference}
An error item is categorized under ``lexical inference'' if the gold label is
inline with the signature of its embedding verb.
Such errors happen on items of a given verb for which the training data do not exhibit a clear pattern because the training items contains items where the verb follows its signature as well as items where pragmatic factors override the signature interpretation. For example, \Next gets a factual interpretation, consistent with the factive signature of \textit{see}.

{\small
\ex. He did \env{not} \cev{see} that Manning had \eventp{glanced}{2}{0.47} at him.

}

However, the training instances with \textit{see} under negation have labels
ranging from $-2$ to $2$ (see the orange \texttimes's in the fourth panel of
Figure~\ref{fig:rp_scatter}).
Some items indeed get a negative label because of the presence of pragmatic factors, such as in \ref{ex:jon}, but the system is unable to identify these factors.
It thus fails to learn to tease apart the factual and nonfactual items, predicting a neutral label that is roughly the mean of the labels of the training items with \textit{see} under negation.

\paragraph{Annotation error} As in all datasets, it seems that some human labels are wrong and the model actually predicts the right label. For instance, \ref{ex:annot_error} should have a more positive label (rather than $0.5$), as
\textit{realize} is taken to be factive and nothing in the context indicates a nonfactual interpretation.

{\small
\ex.\label{ex:annot_error} I did \env{not} \cev{realize} that John had \eventp{fought}{0.5}{2.31} with his mother prior to killing her.

}

In total,  55 items (with absolute errors ranging from 1.10 to 4.35, and a mean of 1.95) were annotated in CB out of 556 items, and 250 in RP (with absolute errors ranging from 1.23 to 4.36, and a mean of 1.70) out of 2,508 items.
Table~\ref{tab:error_types} gives the numbers and percentages of errors in each category.
The two datasets show different patterns that reflect their own
characteristics. CB has rich preceding contexts, and
therefore more items exhibit inferences that can be traced to the effect of context.
RP has more item categorized under lexical inference, because there is not much context to override the default lexical inference.
RP also has more items under annotation errors, due to the limited amount of annotations collected for each item (3 annotations per item).

\begin{table}
\resizebox{\linewidth}{!}{%
\begin{tabular}{lrrcrr}
\toprule
                              & \multicolumn{2}{c}{CB}& & \multicolumn{2}{c}{RP} \\
                                 & \# & \%   && \#& \%    \\
  \midrule
Prior probability of the event   & 5       & 9.1    && 32      & 12.8   \\
Context suggests (non)factuality & 34      & 61.8   && 29      & 11.6   \\
Question Under Discussion (QUD)  &         &        && 20      & 8.0    \\
Tense/aspect                     & 1       & 1.8    && 8       & 3.2    \\
Subject authority/credibility    & 1       & 1.8    && 14      & 5.6    \\
Subject-complement interaction   &         &        && 26      & 10.4   \\
Lexical inference                & 12      & 21.8   && 88      & 35.2   \\
Annotation error                 & 2       & 3.6    && 33      & 13.2   \\
  \midrule
Total items categorized                     & \multicolumn{1}{c}{55}       &&& \multicolumn{1}{c}{250} \\
\bottomrule
\end{tabular}%
}
\caption{
  Numbers (\#) and percentages (\%) of error items categorized for CB and RP.
}
\label{tab:error_types}
\end{table}

Although we only systematically annotated CB and RP (given that these datasets focus on embedded events), the errors in the other datasets focusing on main-clause events also exhibit similar inferences as the ones we categorized above, such as effects of context and lexical inference (more broadly construed).\footnote{Some of the error categories only apply to embedded events, including the effect of Question Under Discussion and subject authority.} Most of the errors concern nominal events. In the following examples, \ref{ex:uw_drop} and \ref{ex:uw_fake} from UW, and \ref{ex:meantime_end} from MEANTIME,
the model failed to take into account the surrounding context which suggests that the events are nonfactual. In \ref{ex:uw_drop}, the lexical meaning of \textit{dropped} clearly indicates that the plan is nonfactual. In \ref{ex:uw_fake}, the death was \textit{faked}, and in \ref{ex:meantime_end} production was \textit{brought to an end}, indicating that the death did not happen and there is no production anymore.

{\small
\ex. \label{ex:uw_drop}
In 2011 , the AAR consortium attempted to block a drilling joint venture in the Arctic between BP and Rosneft through the courts and the \eventp{plan}{-2.8}{1.84} was eventually dropped.

\ex. \label{ex:uw_fake}
The day before Raymond Roth was pulled over, his wife, Evana, showed authorities e-mails she had discovered that appeared to detail a plan between him and his son to fake his \eventp{death}{-2.8}{1.35}

\ex. \label{ex:meantime_end}
Boeing Commercial Airplanes on Tuesday delivered the final 717 jet built to AirTran Airways in ceremonies in Long Beach, California, bringing \eventp{production}{-3}{3.02} of McDonnell Douglas jets to an end.

}

In \ref{ex:factbank_nato}, from FactBank,
\textit{just what NATO will do} carries the implication that NATO will do something, and the \textit{do} event is therefore annotated as factual.

{\small
\ex. \label{ex:factbank_nato}
Just what NATO will \eventp{do}{3}{-0.05} with these eager applicants is not clear.

}

Example \ref{ex:uds} from UDS-IH2 features a specific meaning of the embedding verb \textit{say}: here \textit{say} makes an assumption instead of the usual speech report, and therefore suggests that the embedded event is not factual.

{\small
\ex. \label{ex:uds}
\cev{Say} after I \eventp{finished}{-2.25}{2.38} those 2 years and I found a job.

}

\paragraph{Inter-annotator agreement for categorization}
Both annotators annotated all 55 items in CB. For RP, one of the annotators annotated 190 examples, and the other annotated
100 examples, with 40 annotated by both. Among the set of items that were annotated by both annotators, annotators agreed on the error categorization 90\% of the time for the CB items and 80\% of the time for the RP items.
This is comparable to the agreement level in \citet{DBLP:journals/corr/abs-2010-12729}, in which inferences types for the ANLI dataset \cite{nie-etal-2020-adversarial}
are annotated.

\section{Conclusion}
In this paper, we showed that, although fine-tuning BERT gives strong performance on
several factuality datasets,
it only captures statistical regularities in the data and fails to take into account pragmatic factors which play a role on event factuality.
This aligns with \citet{chaves2020don}'s findings for acceptability of filler-gap dependencies: neural models give the impression that they capture island constraints well when such phenomena can be predicted by surface statistical regularities, but the models do not actually capture the underlying mechanism involving
various semantic and pragmatic factors.
Recent work has found that BERT models have some capacity to
perform pragmatic inferences:
\citet{schuster2019harnessing} for scalar implicatures in naturally occurring data,
\citet{jeretic2020natural} for scalar implicatures and presuppositions
triggered by certain lexical items in constructed data. It is however possible that the good performance on those data is solely driven by surface features as well.
BERT models still only have limited capabilities to account for the wide range of pragmatic inferences in human language.

\section*{Acknowledgment}
We thank TACL editor-in-chief Brian Roark and action editor Benjamin Van Durme for the time they committed to the review process, as well as the anonymous reviewers for their insightful feedback.
We also thank Micha Elsner, Cory Shain, Michael White, and members of the OSU Clippers discussion group for suggestions and feedback.
This material is based upon work supported by the National Science Foundation under Grant No.\ IIS-1845122.

\bibliographystyle{acl_natbib}
\bibliography{main}

\end{document}